\documentclass{article} 
\usepackage{iclr2026_conference,times}

\usepackage{amsmath,amsfonts,bm}

\def\eqref#1{equation~\ref{#1}}

\def\1{\bm{1}}

\DeclareMathAlphabet{\mathsfit}{\encodingdefault}{\sfdefault}{m}{sl}
\SetMathAlphabet{\mathsfit}{bold}{\encodingdefault}{\sfdefault}{bx}{n}

\usepackage{hyperref}
\usepackage{subcaption}
\usepackage{bbding}
\usepackage{mathtools}
\usepackage{amsthm}
\usepackage{url}
\usepackage[table]{xcolor} 
\usepackage{colortbl}      
\usepackage{pifont}        
\usepackage{amsmath}       
\usepackage{amssymb}       
\usepackage{graphicx}      
\usepackage{caption}
\usepackage{booktabs}      
\usepackage{multirow}      
\usepackage{wrapfig}

\usepackage{marvosym}

\definecolor{mygreen}{HTML}{2ECC71} 
\definecolor{myred}{HTML}{E74C3C}     
\definecolor{myblue}{HTML}{3498DB}    

\newcommand{\cmark}{\textcolor{mygreen}{\ding{51}}}
\newcommand{\xmark}{\textcolor{myred}{\ding{55}}}

\newcommand{\uparrowmetric}{\textcolor{mygreen}{\raisebox{-0.2ex}{\scalebox{0.8}{$\uparrow$}}}}
\newcommand{\downarrowmetric}{\textcolor{mygreen}{\raisebox{-0.2ex}{\scalebox{0.8}{$\downarrow$}}}}

\newcommand{\ie}{\emph{i.e., }}
\newcommand{\eg}{\emph{e.g., }}

\title{IGGT: Instance-Grounded Geometry Transformer for Semantic 3D Reconstruction}

\author{
Hao Li$^{1,2,3}$,
Zhengyu Zou$^{1}$,
Fangfu Liu$^{4}$,
Xuanyang Zhang$^{3}$$^\ast$, 
Fangzhou Hong$^{2}$,  \\
\textbf{Yukang Cao}$^{2}$,
\textbf{Yushi Lan}$^{2}$,
\textbf{Manyuan Zhang}$^{5}$,
\textbf{Gang Yu}$^{3}$,
\textbf{Dingwen Zhang}$^{1}$\textsuperscript{\Letter},
\textbf{Ziwei Liu}$^{2}$
\\
$^1$NWPU \quad
$^2$S-Lab, NTU \quad
$^3$StepFun, Inc. \quad
$^4$THU \quad
$^5$MMLab, CUHK
}

\iclrfinalcopy
\begin{document}
\maketitle

{
  \renewcommand{\thefootnote}%
    {}
  \footnotetext[2]{ $^\ast$Project Leader. \textsuperscript{\Letter}Corresponding Authors.}
 
}
\begin{center}
    \centering
    \vspace{-20pt}
    \includegraphics[width=0.95\linewidth]{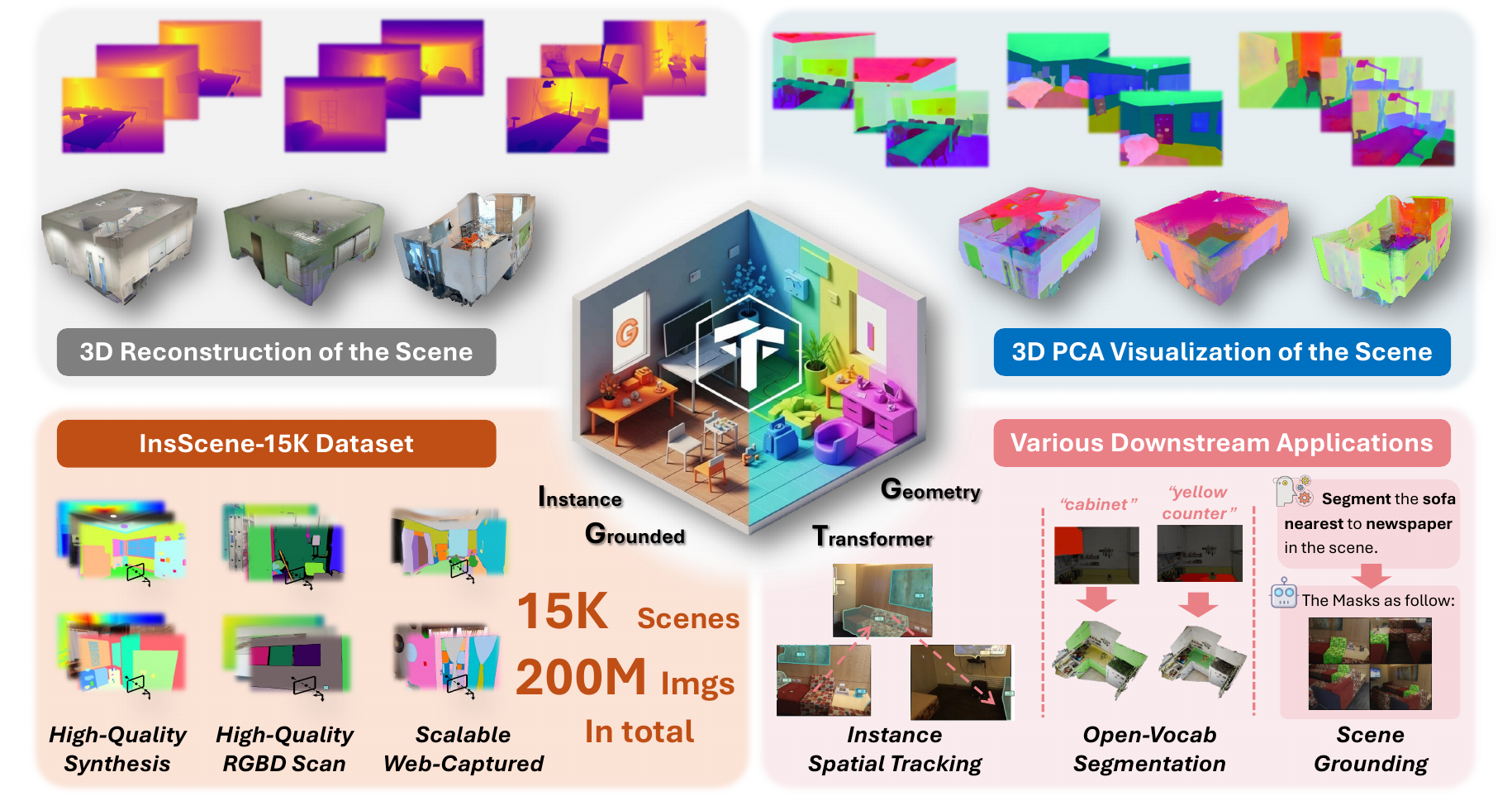}
    \captionof{figure}{\textbf{IGGT}: building upon our curated large-scale dataset InsScene-15K, we propose a novel end-to-end framework that enables geometric reconstruction and contextual understanding in a unified representation. This paradigm facilitates a wide range of applications, including spatial tracking, 2D / 3D open-vocabulary segmentation, and scene grounding.}
    \label{fig:teaser}
\end{center}

\begin{abstract}
Humans naturally perceive the geometric structure and semantic content of a 3D world as intertwined dimensions, enabling coherent and accurate understanding of complex scenes.
However, most prior approaches prioritize training large geometry models for low-level 3D reconstruction and treat high-level spatial understanding in isolation, overlooking the crucial interplay between these two fundamental aspects of 3D-scene analysis, thereby limiting generalization and leading to poor performance in downstream 3D understanding tasks. 
Recent attempts have mitigated this issue by simply aligning 3D models with specific language models, thus restricting perception to the aligned model’s capacity and limiting adaptability to downstream tasks.
In this paper, we propose \textbf{Instance-Grounded Geometry Transformer (IGGT)}, an end-to-end large unified transformer to unify the knowledge for both spatial reconstruction and instance-level contextual understanding. Specifically, we design a \textit{3D-Consistent Contrastive Learning} strategy that guides IGGT to encode a unified representation with geometric structures and instance-grounded clustering through only 2D visual inputs. This representation supports consistent lifting of 2D visual inputs into a coherent 3D scene with explicitly distinct object instances. To facilitate this task, we further construct \textbf{InsScene-15K}, a large-scale dataset with high-quality RGB images, poses, depth maps, and 3D-consistent instance-level mask annotations with a novel data curation pipeline. 
Unlike previous methods that bound with a specific language model, we introduce an \textit{Instance-Grounded Scene Understanding} paradigm, where instance masks serve as the bridge connecting our unified representation with diverse Visual Language Models (VLMs) in a plug-and-play manner, substantially expanding downstream understanding capabilities.
Extensive experiments on instance spatial tracking, open-vocabulary segmentation, and QA scene grounding demonstrate that IGGT outperforms state-of-the-art methods in both quality and consistency for semantic 3D reconstruction. \url{https://github.com/lifuguan/IGGT_official}.

\end{abstract}

\section{Introduction} 
A foundational goal in the pursuit of spatial intelligence~\citep{yang2025thinking} is to build representations that mirror human understanding—capturing both the precise geometric structure and rich semantic content of a scene from visual sensory inputs such as RGB images.
Such representations are vital for enabling downstream tasks like robotic manipulation~\citep{qu2025spatialvla}, AR / VR~\citep{jiang2025anysplat}, and planning~\citep{zhang2024vision}. 

Previous methods~\citep{zust2025panst3r,fan2024large-lsm,sun2025uni3r} tackle this challenge through a fragmented paradigm, decoupling 3D geometric reconstruction and high-level semantic understanding into isolated tasks. Typically, they first leverage geometry-focused techniques (\eg Multi-View Stereo (MVS) methods~\citep{schoenberger2016mvs, schoenberger2016sfm} or off-the-shelf large Image-to-3D models~\citep{wang2024dust3r, wang2025vggt}) to predict low-level 3D structures, followed by vision-language models (VLMs)~\citep{bai2023qwen, bai2025qwen2} or 2D segmentation models~\citep{cheng2022masked} to perform high-level semantic segmentation tasks. However, these disjointed approaches are inherently flawed, as they propagate errors between stages and fail to leverage the mutual context between shape and identity, preventing them from enhancing each other's capabilities and hindering their ability to support model reconstruction. 

Recently emerged methods~\citep{fan2024large-lsm, sun2025uni3r} attempt to bridge this gap by aligning spatial models with specific VLM~\citep{lseg}. 
However, these approaches suffer from three critical limitations. 
First, since 3D geometry contains low-level, fine-grained structural signals, forcing a strict alignment with high-level textual concepts can over-smooth the representation, degrading high-frequency geometric details and undermining multi-view consistency. 
Second, this tight coupling to a specific VLM architecture inherently restricts the performance to the base model (\textit{e.g.}, LSeg~\citep{lseg}) and prevents the integration of newer, more powerful foundation models (\textit{e.g.}, CLIP~\citep{radford2021clip}, SigLIP~\citep{tschannen2025siglip}).
Third, since these VLMs~\citep{lseg,ghiasi2022scaling} are mainly trained on 2D image–text pairs, their aligned features often fail to distinguish objects within the same semantic category, which significantly limits more downstream applications (\eg, 3D instance-consistent tracking under large viewpoint changes and spatial QA when interfaced with VLMs). 


To address this, we propose \textbf{I}nstance-\textbf{G}rounded \textbf{G}eometry \textbf{T}ransformer (\textbf{IGGT}), a novel end-to-end framework that unifies the representation for spatial reconstruction and contextual understanding. Instead of simply aligning geometry with language features, our key idea is to \textit{couple both factors by joint training and encourage the model to autonomously learn the relationship between 3D instance-level semantics and their geometric structures, yielding \textbf{mutual improvements} in contextual understanding and geometry reconstruction}. 
Specifically, \textbf{1)} we employ a large Unified Transformer to encode multi-view images into unified token representations of the 3D scene, which are decoded by a Geometry Head and an Instance Head into geometric point maps and an instance clustering fields, respectively.
\textbf{2)} we employ a cross-modal fusion block with a window-shifted attention mechanism, enabling the Instance Head to leverage fine-grained geometric features at pixel level to enhance its spatial awareness.
\textbf{3)} To further improve multi-view consistency of the instance fields, we design a 3D-consistent contrastive learning strategy that guides IGGT to learn both geometric structures and instance-grounded clustering features.
As instance-level geometry-semantics aligned annotations remain scarce in the community, we facilitate this task by presenting a large-scale dataset coined \textbf{InsScene-15K}, a meticulously constructed dataset comprising high-quality RGB images, poses, depth maps, and 3D-consistent instance masks.

\textbf{One more thing}, after training the full model (\ie IGGT), we design an \textit{Instance-Grounded Scene Understanding} strategy, where instance masks serve
as the bridge connecting IGGT with diverse VLMs.
Such a paradigm not only enables the seamless, plug-and-play integration of various vision-language models (VLMs) such as CLIP and SigLIP to lift downstream task performance, but also extends to Large Multimodal Models (LMMs)~\citep{bai2023qwen,bai2025qwen2}, unlocking more sophisticated scene understanding and a broader spectrum of applications like scene grounding. 

We validate our framework through extensive experiments on diverse downstream tasks (\eg spatial tracking segmentation, open-vocabulary segmentation, and scene grounding), demonstrating its superiority over state-of-the-art methods in both task performance and 3D scene coherence.

\section{InsScene-15K Dataset}
We construct the InsScene-15K dataset (in Sec.~\ref{sec:dataset}), where each scene includes corresponding RGB images, depth maps, poses, and 3D-consistent instance segmentation masks. To maintain consistency, we ensure that each instance retains a unique ID across all views.
\begin{figure}[t]
    \centering
    \includegraphics[width=0.9\linewidth]{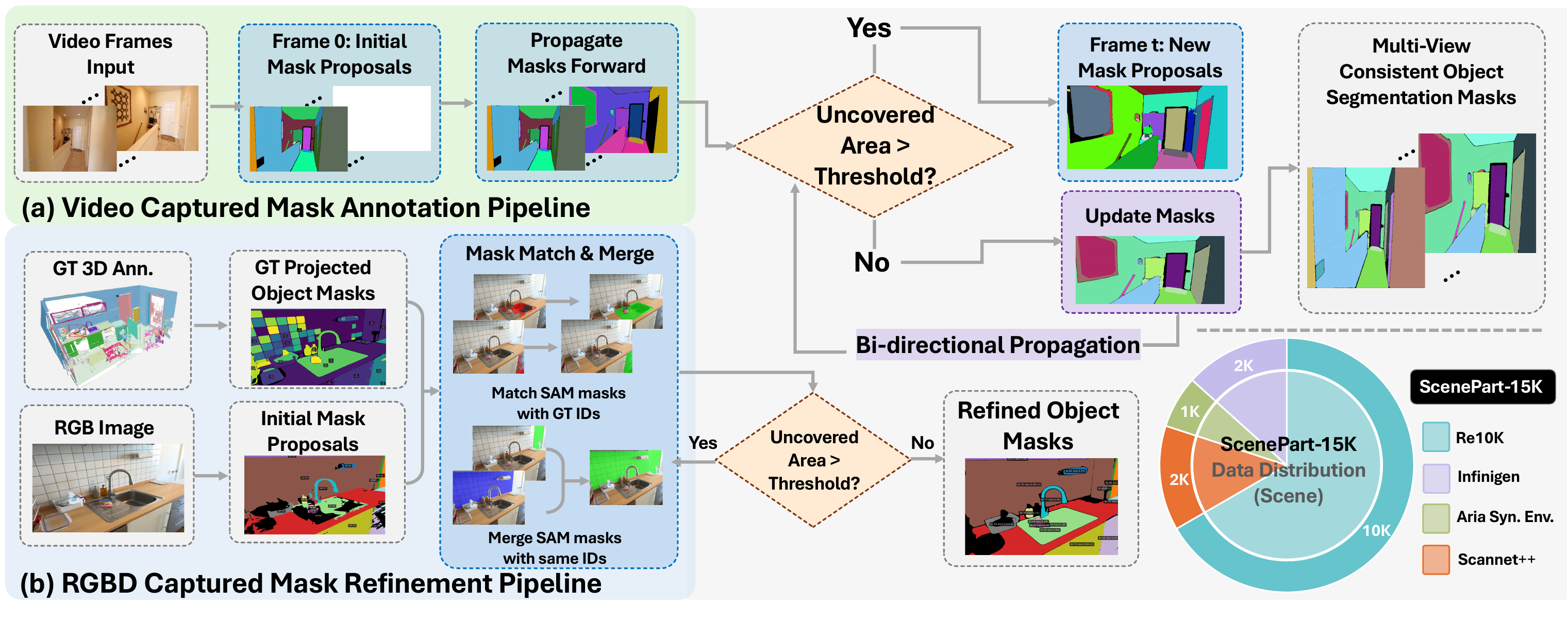}
    \caption{\textbf{Data Curation Pipeline.} Our data is collected from various sources and then annotated by a novel data engine driven by SAM2~\citep{ravi2024sam2}. (a) For video captured scenes (\textit{i.e.}, RE10k~\citep{zhou2018stereo}), we annotate them through a customized SAM2 video dense prediction pipeline. (b) For RGBD-scan scenes (e.g., ScanNet++~\citep{yeshwanth2023scannet++}), we regenerate dense mask annotations for each image and align them with the projected coarse GT masks.
    }
\vspace{-0.2in}
    \label{fig:dataset curation}
\end{figure}
\begin{figure}[b]
\vspace{-0.2in}
    \centering
    \includegraphics[width=0.9\linewidth]{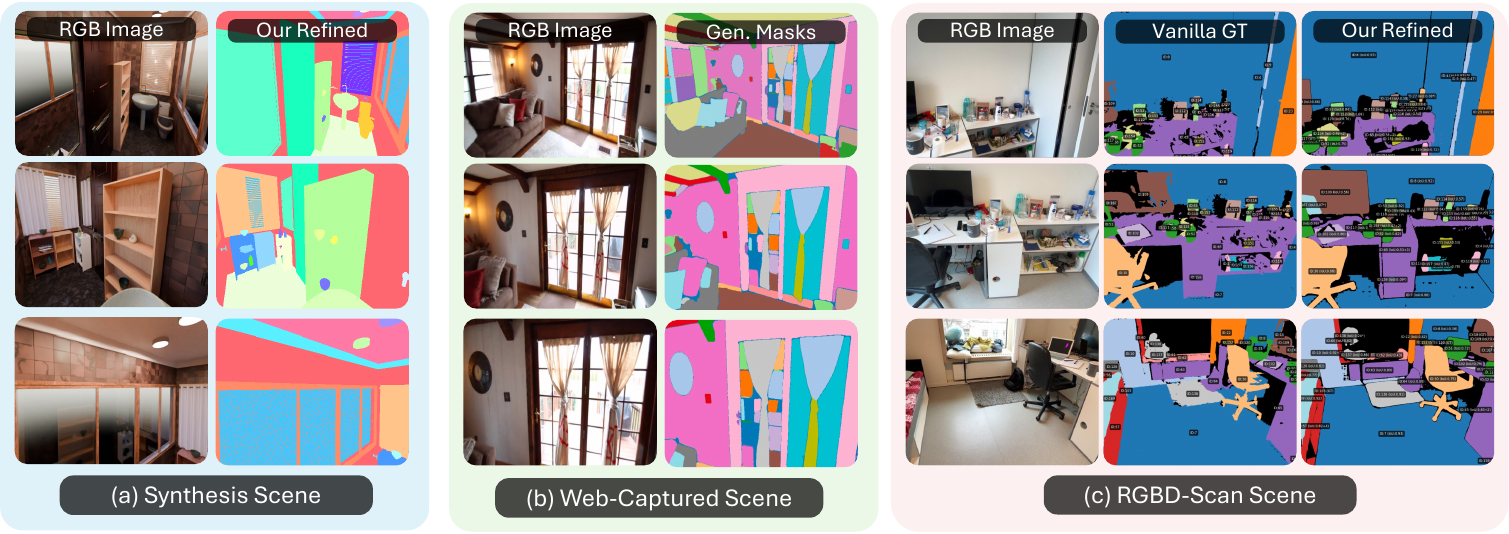}
    \caption{Visualization of mask annotations from three different sources. For the RGBD-scan scene, we additionally compare the vanilla ground-truth masks from ScanNet++~\citep{yeshwanth2023scannet++} with our refined annotations, along with their corresponding matched IDs and mIoU scores.}
    \label{fig:anno_vis}
\end{figure}
\label{sec:dataset}

Our data curation pipeline systematically integrates three distinct categories of data to ensure comprehensiveness and diversity, as illustrated in Fig.~\ref{fig:dataset curation}: 1) synthesis (Aria~\citep{pan2023aria}, Infinigen~\citep{infinigen2024indoors}); 2) Video captured (RE10K~\citep{zhou2018stereo}); 3) RGBD captured (Scannet++~\citep{yeshwanth2023scannet++}).
For synthetic datasets (\textit{e.g.}, Aria and Infinigen), we simultaneously generate the RGB image, depth map, camera pose, and object-level segmentation masks for each rendered view. Since the simulation environment provides perfectly accurate 2D ground-truth masks (in Fig.~\ref{fig:anno_vis} (a)), we use them directly without any post-processing.
Moreover, regarding real-world scenarios, we propose a novel data curation pipeline that includes multi-view mask annotation and refinement stages, driven by SAM2~\citep{ravi2024sam2}.
Specifically, for real-world video-captured scenes such as RE10K~\citep{zhou2018stereo} (Fig.~\ref{fig:dataset curation}a), our method first employs SAM to generate dense mask proposals on the initial frame. These proposals are then used as prompts for the SAM2 video object segmenter to propagate masks temporally throughout the sequence. To handle new objects and mitigate drift, we adopt an iterative strategy that designates a new keyframe whenever the unsegmented area increases, where SAM is reapplied to discover objects in the uncovered regions. After processing the entire video, a final bi-directional propagation pass ensures high temporal consistency across object tracks. This curation strategy provides scalable and diverse annotations that enhance the generalization ability of our model.

For challenging datasets with large-scale camera motion but coarse 3D annotations such as ScanNet++~\citep{yeshwanth2023scannet++}, we first project the 3D annotations into 2D to obtain initial image-level object masks. While this guarantees multi-view consistency of object IDs, the masks are often coarse and imprecise. To improve their quality, we use SAM2 to generate fine-grained initial mask proposals that are accurate in shape but lack identity information. These proposals are then aligned with the projected ground-truth masks to assign consistent object IDs (Fig.~\ref{fig:anno_vis} (c)), and proposals belonging to the same ID are merged into complete masks. The process is iteratively refined until all image regions are covered. This pipeline (Fig.~\ref{fig:dataset curation} (b)) achieves both multi-view ID consistency and shape-accurate annotations, substantially improving 2D mask quality for real-world scenarios.

\section{Methodology}
\subsection{Overview}
Our method consists of two main phases. 
Firstly, we propose IGGT (in Sec.~\ref{sec:IGGT}), a unified foundation model that simultaneously predicts instance-discriminative features at the spatial level and performs 3D reconstruction through 3D-consistent contrastive learning on large-scale datasets.
%
Secondly, we propose an instance-grounded scene understanding strategy (Sec.~\ref{sec:post}). This strategy employs unsupervised clustering to partition the scene into instances by grouping the predicted features into masks with consistent instance IDs. These masks are then used to guide state-of-the-art vision-language models (VLMs, e.g., CLIP, OpenSeg) and large multimodal models (LMMs, e.g., GPT-4o, Qwen2.5-VL) to perform open-vocabulary scene querying and grounding tasks.

\subsection{Architecture of IGGT}
\label{sec:IGGT}
\begin{figure}
    \centering
    \includegraphics[width=0.9\linewidth]{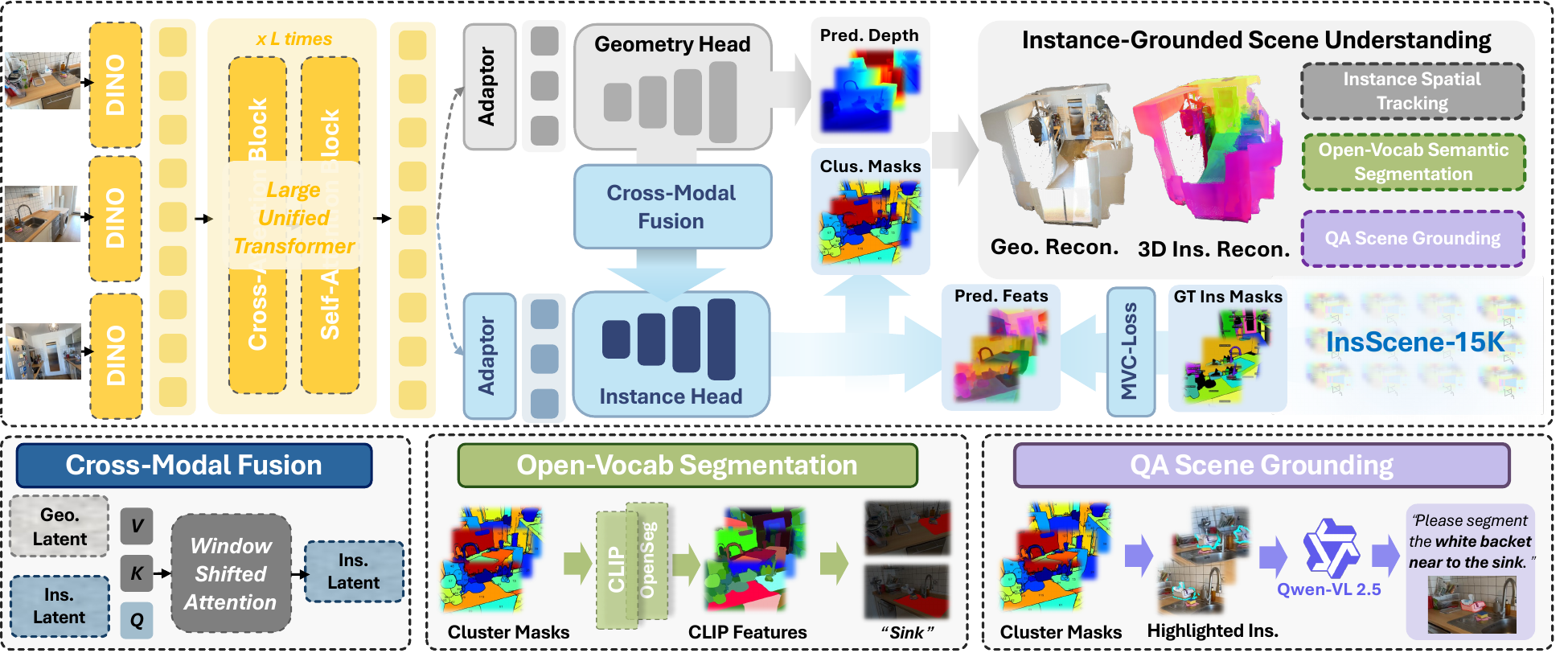}
    \caption{Overview of \textbf{IGGT}. Given input images, our method encodes them into a series of Unified Token Representations, which are then processed by the Geometry Head and the Instance Head to produce high-quality geometric reconstructions and instance-grounded clusterings simultaneously. In the end, we introduce Instance-Grounded Scene Understanding to perform multiple applications.}
    \label{fig:framework}
    \vspace{-0.2in}
\end{figure}

As illustrated in Fig.~\ref{fig:framework}, given \(N\) input images \(\{I^i\in\mathbb{R}^{H \times W \times 3}\}^{N}_{i=1}\), we aim to forge a unified representation, enabling comprehensive 3D reconstruction and understanding in a mutually reinforcing manner. Specifically, we propose IGGT \(\mathcal{F}\), which predicts camera parameters \(t_i\), depth map \(D_i\), point map \(P_i\), and 3D-consistent, instance-level feature maps \(S_i\) in a feed-forward manner:
{
\small
\begin{equation}
    \mathcal{F} : \{I_i\}_{i=1}^N \mapsto (t_i, D_i, P_i, S_i)_{i=1}^N.
\end{equation}
}Our IGGT consists of three parts: 1) a Large Unified Transformer to capture Unified Token Representation from multiple images; 2) two Downstream Heads with a Cross-Modal Fusion Block to simultaneously predict geometric structures and corresponding instance features via a mutual enhancement pattern; 3) a 3D consistent supervision to empower the model to construct 3D-consistent feature fields.

\noindent \textbf{Large Unified Transformer.}
We follow VGGT to construct a 1B parameter large unified Transformer, designed to encode the multi-view images \(\{I_i\}_{i=1}^N\) into a set of powerful unified token representations \(\{\mathbf{T}_i\in\mathbb{R}^{M\times D}\}_{i=1}^N\), where \(M\) denotes the numbers of the tokens for each image and \(D\) is the dimension of the token.
Our large Unified Transformer first adopts pretrained DINOv2~\citep{oquab2023dinov2} to extract patch-level image tokens. To support arbitrary multi-view inputs while maintaining permutation equivariance, a learnable camera token is concatenated to each view's token sequences.  
Subsequently, 24 blocks of intra-view self-attention and global-view cross-attention are applied to transform the image tokens into unified tokens $\{\mathbf{T}_i\}_{i=1}^N$, capturing both local and global context, which enables a holistic and globally consistent understanding of the 3D scene.

\noindent \textbf{Downstream Heads and Cross-Modal Fusion Block.} We employ two downstream branches—Geometry Head and Instance Head—to decode the unified tokens \(\mathbf{T}_i\}\) into geometric and instance features, respectively.
The Geometry Head, inheriting its design from VGGT, is composed of three distinct modules: a camera predictor, a depth predictor, and a point predictor. The camera predictor is tasked with regressing camera parameters, including extrinsics and intrinsics, from camera-specific tokens.
For dense prediction, the depth and point predictors employ a DPT-like architecture~\citep {ranftl2021vision}. This architecture reconstructs a hierarchical geometric feature \(\boldsymbol{F}^{pt}_i = \{ F_{i,(l)}^{pt} \}_{l=1}^{4}\) from the unified tokens through progressive upsampling and multi-scale fusion network \(\Phi_{pt}(\cdot)\).
Similar to this dense prediction paradigm, our Instance Head \(\Phi_{ins}(\cdot)\) also adopts a DPT-like architecture to perform dense instance features \(\boldsymbol{F}^{ins}_i = \{ F_{i,(l)}^{ins} \}_{l=1}^{4}\):
{
\small
\begin{equation}
    \{{F}^{pt}_i\} = \Phi_{pt}(\{\mathbf{T}_i\}), \quad \{{F}^{ins}_i\} = \Phi_{ins}(\{\mathbf{T}_i\}).
\end{equation}
}Moreover, to enhance the fine-grained spatial awareness of the instance head, we propose a cross-modal fusion block \(\mathcal{F}_{\text{win}}(\cdot)\), which utilizes a sliding window cross attention to embed spatial structure into the instance representation, making them more sensitive to object boundaries and spatial layouts while avoiding the quadratic complexity of global attention:
{\small
\begin{equation}
\hat{F}_{i,(l)}^{ins} = F_{i,(l)}^{ins}+\mathcal{F}_{\text{win}}(Q=F_{i,(l)}^{ins}, K=F_{i,(l)}^{pt}, V=F_{i,(l)}^{pt}).
\end{equation}
}After that, we concatenate all refined instance features \(\{\hat{F}_{i,(l)}^{ins}\}\) and map them through a conventional \(3\times 3\) convolutional layer to 8 dimensional instance features \(O_{ins}\in\mathbb{R}^{N\times 8\times H\times W}\).

\noindent \textbf{3D-Consistent Contrastive Supervision.} We enforce 3D consistency on the instance features $O_{\text{ins}} \in \mathbb{R}^{N \times 8 \times H \times W}$ by applying a multi-view contrastive loss $\mathcal{L}_{mvc}$, which is designed to pull features from the same 3D instance together across views while pushing features from different instances apart. Given a set of sampled pixels $\mathcal{P}$, the loss is formulated as:
{\small
\begin{equation}
\mathcal{L}_{mvc} = \lambda_{pull} \cdot  \sum_{\substack{p_i, p_j \in \mathcal{P} \\ m(p_i)=m(p_j)}} d(f_{p_i}, f_{p_j}) + \lambda_{push} \cdot \sum_{\substack{p_i, p_j \in \mathcal{P} \\ m(p_i) \neq m(p_j)}} \max(0, M - d(f_{p_i}, f_{p_j}))
\end{equation}
}Here, $d(\cdot, \cdot)$ is the L2 distance between normalized features, $m(p_i)$ is the instance ID of pixel $p_i$. The coefficients $\lambda_{\text{pull}}$ and $\lambda_{\text{push}}$ balance the pulling and pushing terms, while $M$ is a margin hyperparameter that controls the discriminative between different instances. 
This objective structures the instance representations according to the 3D scene geometry, improving generalization. Overall, we train the whole model in a multi-task loss:
{\small
\begin{equation}
    \mathcal{L}_{overall} = \mathcal{L}_{pose} + \mathcal{L}_{depth} + \mathcal{L}_{pmap} + \mathcal{L}_{mvc},
\end{equation}
}where geometry supervision terms pose \(\mathcal{L}_{pose}\), depth \(\mathcal{L}_{depth}\), and point map \(\mathcal{L}_{pmap}\) are followed by the training paradigm of VGGT, which is used to supervise the outputs of the geometry head.

\subsection{Instance-Grounded Scene Understanding}
\label{sec:post}
Unlike prior approaches that are tightly coupled with a specific language model (e.g., for Open-Vocabulary Segmentation) and thus limited to a single type of task, we decouple our framework from specific language models and propose a novel Instance-Grounded Scene Understanding strategy to support a broad range of downstream tasks. As shown in Tab.~\ref{tab:scannet}, our method is the only one that simultaneously enables spatial tracking, image-to-3D reconstruction, and scene understanding, while achieving state-of-the-art performance across all tasks.

\noindent \textbf{Instance Spatial Tracking.}
Specifically, inspired by SAMPart3D~\citep{liu2025partfield}, we apply the density-based clustering algorithm HDBSCAN~\citep{mcinnes2017hdbscan} that gathers multi-view 2D instance features \(\{O^{ins}_i\}\) into \(K\) distinct clusters, where each cluster represents a unique object instance present in the scene. Then we re-project the assigned cluster labels to their corresponding pixel locations produces a set of 3D-consistent 2D instance masks \(\{M^{ins}_{i,k}\}^{K}_{k=1}\). Such a paradigm enables dense tracking and segmentation of specific instances across multi-view images by leveraging explicit 3D priors, in stark contrast to existing methods that are either limited to discriminating category-level features or lose targets during significant camera motion.

\noindent \textbf{Open-Vocabulary Semantic Segmentation.}
These 3D-consistent instance masks serve as effective prompts for any off-the-shelf VLMs~\citep{radford2021clip, ghiasi2022scaling}, enabling them to perform robust open-vocabulary semantic segmentation by assigning a semantic category to each mask-defined region. Here we take OpenSeg~\citep{ghiasi2022scaling} as an example. It first produces image-wise features \(\{F_i^{lang}\in\mathbb{R}^{D\times H \times W}\}^N_{i=1}\), which considers contextual information to enable accurate visual-language alignment of the features. 
We then aggregate the features within each 2D instance mask \(\{\textbf{f}_k^{lang}\in\mathbb{R}^{D}\}^K_{i=1}\) via average mask pooling, yielding a compact representation for each instance. This step not only integrates the mask priors into the visual-language space, but also sharpens object boundaries and captures fine-grained local category cues, making the subsequent semantic assignment more accurate and robust.

\noindent \textbf{QA Scene Grounding.}
Unlike prior methods that directly align 3D features with language embeddings, our approach offers greater flexibility by decoupling instance clusterings, which can then interact with LMMs~\citep{bai2025qwen2,team2023gemini} to support object-centric QA in 3D scenes. Concretely, as shown in Fig.~\ref{fig:framework}, given $N$ views, we highlight the image regions corresponding to the same instance $k$ with masks $\{M^{\text{ins}}_{i,k}\}_{i=1}^{N}$ (rendered in red), and query the LMM with yes/no questions to verify object consistency across views. Finally, we aggregate all positive (“yes”) responses and concatenate the corresponding masks to form the final segmentation output.

\begin{table}[t]
\centering
\caption{Quantitative Results on Scannet~\citep{dai2017scannet}. Here we showcase the capability overview and report the spatial track quality, reconstruction accuracy, and 2D / 3D open-vocabulary semantic segmentation accuracy. The \textbf{bold} denotes the best results.}
\label{tab:scannet}
\setlength{\tabcolsep}{2pt}
\resizebox{\textwidth}{!}{
\arrayrulecolor{black}
\begin{tabular}{l|ccc|cc|cc|ccc}
\toprule
\multicolumn{1}{c|}{\textbf{Model}} & \multicolumn{3}{c|}{\textbf{Capability}}  & \multicolumn{2}{c|}{\textbf{Spatial Track}} & \multicolumn{2}{c|}{\textbf{Recon. Metric}} & \multicolumn{3}{c}{\textbf{Open-Vocab. Semantic Segment}} \\
\cmidrule(lr){2-4} \cmidrule(lr){5-6} \cmidrule(lr){7-8} \cmidrule(lr){9-11}
& Recon. & Understand & Track & T-mIoU\uparrowmetric & T-SR\uparrowmetric & Abs. Rel\downarrowmetric & $\tau$\uparrowmetric & 2D mIoU\uparrowmetric & 2D mAcc\uparrowmetric & 3D mIoU\uparrowmetric \\
\midrule
LSeg & \xmark & \cmark & \xmark & - & - & - & - & 58.11& 65.76 & - \\
OpenSeg & \xmark & \cmark & \xmark & - & - & - & - & 42.33& 68.06& - \\
NeRF-DFF & \cmark & \cmark & \xmark & - & - & 7.99 & 36.53 & 45.40 & 65.29 & 12.29\\
Feature-3DGS & \cmark & \cmark & \xmark & - & - & 6.48 & 41.63 & 57.69& 63.26& 23.42\\
LSM (2 Views) & \cmark & \cmark & \xmark  & - & - & 4.22 & 58.65 & 53.07& 53.86& - \\
LSM (Multi-Views) & \cmark & \cmark & \xmark  & - & - & 3.17 & 64.81 & 53.40& 59.50& 35.37\\
SpaTracker+SAM & \xmark & \xmark & \cmark & 26.43 & 38.57 & - & - & - & - & - \\
SAM2* & \xmark & \cmark & \cmark & 53.74 & 71.25 & - & - & - & - & - \\
VGGT & \cmark & \xmark & \xmark  & - & - & \textbf{1.84} & 83.60 & - & - & - \\
\rowcolor{myblue!25} 
Ours& \cmark & \cmark & \cmark  & \textbf{69.41} & \textbf{98.66} & 1.90 & \textbf{83.71} & \textbf{60.46}& \textbf{81.84}& \textbf{39.68}\\
\bottomrule
\end{tabular}
\vspace{-0.2in}
}
\end{table}

\begin{table}[t]
\vspace{-0.8em}
\centering
\caption{Quantitative Results on Scannet++~\citep{yeshwanth2023scannet++}. Here we report the spatial track quality, reconstruction accuracy, and 2D / 3D open-vocabulary semantic segmentation accuracy.}
\label{tab:scannet++}
\resizebox{0.8\columnwidth}{!}{
\begin{tabular}{l|cc|cc|ccc}
\toprule
\multicolumn{1}{c|}{\textbf{Model}} & \multicolumn{2}{c|}{\textbf{Spatial Track}} & \multicolumn{2}{c|}{\textbf{Recon. Metric}} & \multicolumn{3}{c}{\textbf{Open-Vocab. Semantic Segment}} \\
\cmidrule(lr){2-3} \cmidrule(lr){4-5} \cmidrule(lr){6-8}
& T-mIoU\uparrowmetric & T-SR\uparrowmetric & Abs. Rel\downarrowmetric & $\tau$\uparrowmetric & 2D mIoU\uparrowmetric & 2D mAcc\uparrowmetric & 3D mIoU\uparrowmetric \\
\midrule
LSeg & - & - & - & - & 22.61& 34.42& - \\
OpenSeg & - & - & - & - & 13.92& 48.13& - \\
Feature-3DGS & - & - & 5.92 & 41.64 & 22.47& 33.14& 10.59\\
LSM (2 Views) & - & - & 4.22 & 74.02 & 17.76& 26.95& - \\
LSM (Multi-Views) & - & - & 2.96 & 83.28 & 17.88& 27.84& 15.17\\
SpaTracker+SAM & 16.15 & 23.68 & - & - & - & - & - \\
SAM2* & 44.16 & 57.89 & - & - & - & - & - \\
VGGT & - & - & 2.75 & 85.41 & - & - & - \\
\rowcolor{myblue!25} 
Ours & \textbf{73.02} & \textbf{98.90} & \textbf{2.61} & \textbf{85.66} & \textbf{31.31}& \textbf{70.78}& \textbf{20.14}\\
\bottomrule
\end{tabular}
}
\vspace{-0.2in}
\end{table}

\section{Experiments}

\begin{figure}[t]
    \centering
    \includegraphics[width=0.9\linewidth]{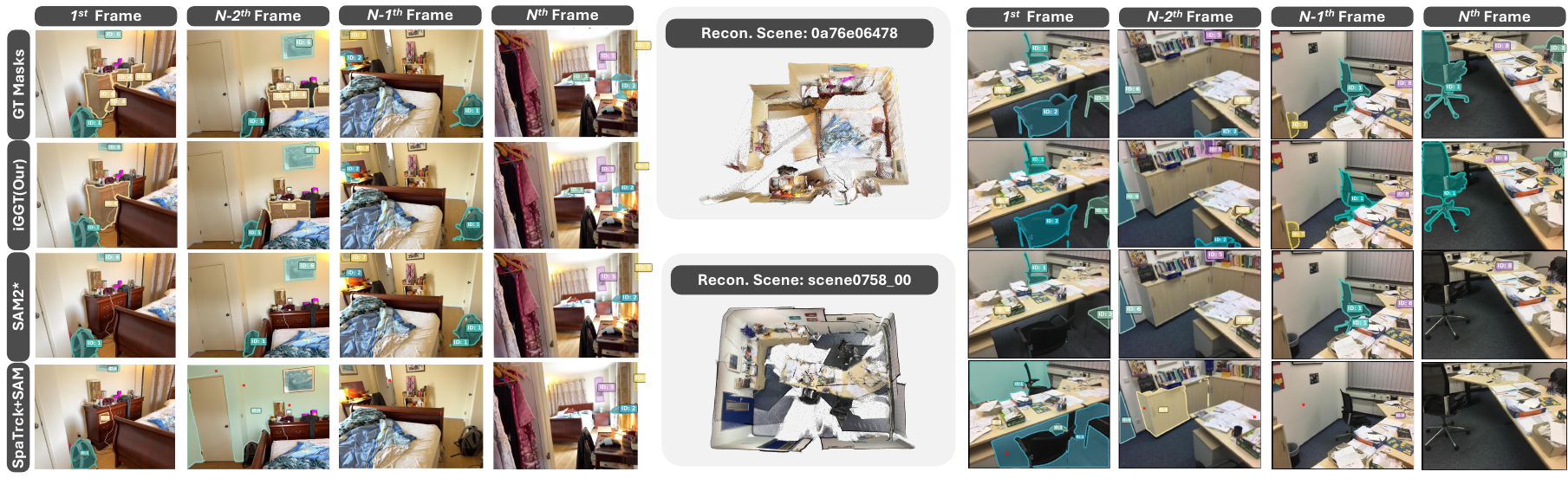}
    \caption{Qualitative results on Instance Spatial Tracking. We present two example scenes from ScanNet~\citep{dai2017scannet} and ScanNet++~\citep{yeshwanth2023scannet++}, and compare our method with SAM2* and SpaTracker+SAM. All instances are visualized with distinct IDs and colors for clarity.}
\vspace{-0.2in}
    \label{fig:vis_3d_tracking}
\end{figure}

\noindent \textbf{Evaluation Details.}
We conduct comprehensive experiments on the ScanNet~\citep{dai2017scannet} and ScanNet++~\citep{yeshwanth2023scannet++} datasets. From each dataset, we randomly select 10 scenes and sample 8-10 images per scene, with the selection strategy designed to maximize spatial coverage of the scene while preserving sufficient overlap to ensure cross-view consistency. (a) For \textit{Instance Spatial Tracking} evaluation, we evaluate tracking performance using Temporal mIoU (T-mIoU) and Temporal Success Rate (T-SR). T-mIoU measures the segmentation accuracy of the same object across different views, while T-SR assesses whether the object is successfully tracked in every view. (b) For \textit{Open-Vocabulary Segmentation} evaluation, we follow LangSplat~\citep{qin2024langsplat} and LangSurf~\citep{li2024langsurf}, which adopt mIoU and mAcc to measure 2D segmentation accuracy. In addition, we evaluate the 3D mIoU metric by aligning the reconstructed scene with the ground-truth point cloud. (c) For \textit{Reconstruction} evaluation, we follow LSM~\citep{fan2024large-lsm} and VGGT~\citep{wang2025vggt} that utilize Absolute Relative Error (Abs. Rel) and Inlier Ratio (\(\tau\)) with a threshold of 1.03 to assess each scene.
The details of these metrics are shown in the appendix.

\begin{figure}[t]
    \centering
    \includegraphics[width=0.9\linewidth]{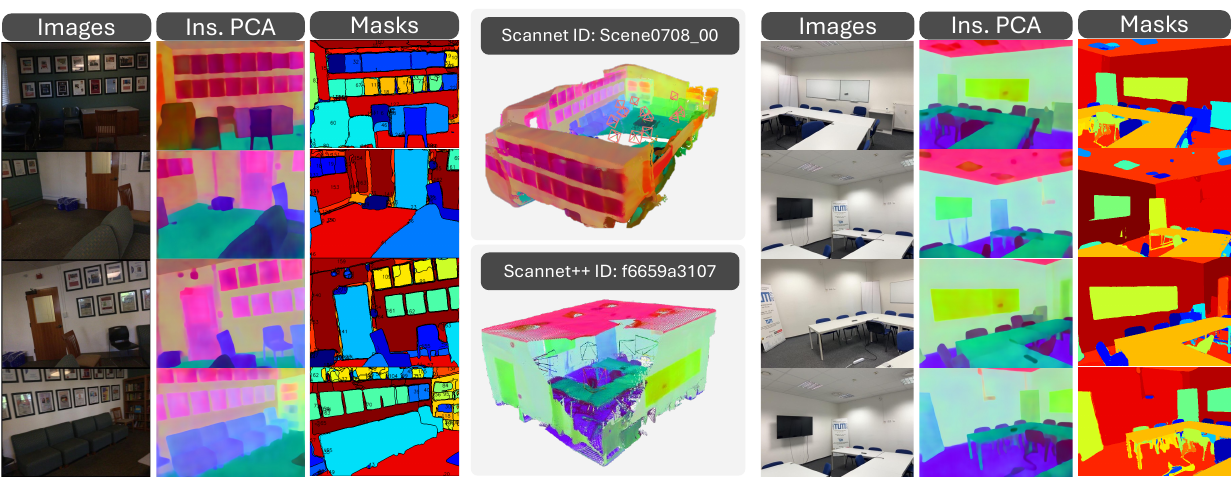}
        \vspace{-0.1in}
    \caption{We visualize our 3D-consistent PCA results with corresponding clustered masks derived from instance-grounded features. Similar colors in PCA indicate higher feature similarity between instances. For clustered masks, the same object instance shares the same color across multi-views.}
        \vspace{-0.2in}
    \label{fig:pca_mask_vis}
\end{figure}

\noindent \textbf{Evaluation of Instance Spatial Tracking.}
To comprehensively evaluate the tracking quality of our proposed method and competing approaches, particularly under large viewpoint changes with multiple objects, we manually annotate a subset of objects across several scenes with precise ground-truth labels (more visualization in the Appendix). For baseline methods, we modify SAM2~\citep{ravi2024sam2} to support dense segmentation and tracking under multi-view inputs, denoted as SAM2*. In addition, we integrate SAM into SpaTrackerV2~\citep{xiao2025spatialtrackerv2}, where tracking points are used as prompts to perform dense segmentation. 
Tab.~\ref{tab:scannet} and Tab.~\ref{tab:scannet++} present the quantitative results, demonstrating the significant superiority of our method. By leveraging implicit 3D reasoning, our approach successfully distinguishes object identities to achieve nearly 100\% T-SR accuracy. In contrast, baseline methods fail at this crucial task, yielding a T-mIoU below 30\%, whereas our approach surpasses 60\%. This performance gap is visually demonstrated in Fig.~\ref{fig:vis_3d_tracking}, where our method successfully tracks and segments the chair under large camera motions, while competing methods lose the track.

Furthermore, we provide additional visualizations of our 3D-consistent instance features using Principal Component Analysis (PCA), along with their corresponding clustered masks, as shown in Fig.~\ref{fig:pca_mask_vis}. As illustrated, IGGT produces 3D-consistent instance-grounded features that remain discriminative across multiple views: multiple instances with the same category exhibit similar yet distinguishable colors in the PCA space. This property serves as a crucial foundation for the Instance Spatial Tracking task, as it enables consistent tracking and segmentation of individual objects even under large motions and in the presence of many similar instances.

\begin{figure}
        \centering
        \includegraphics[width=0.9\linewidth]{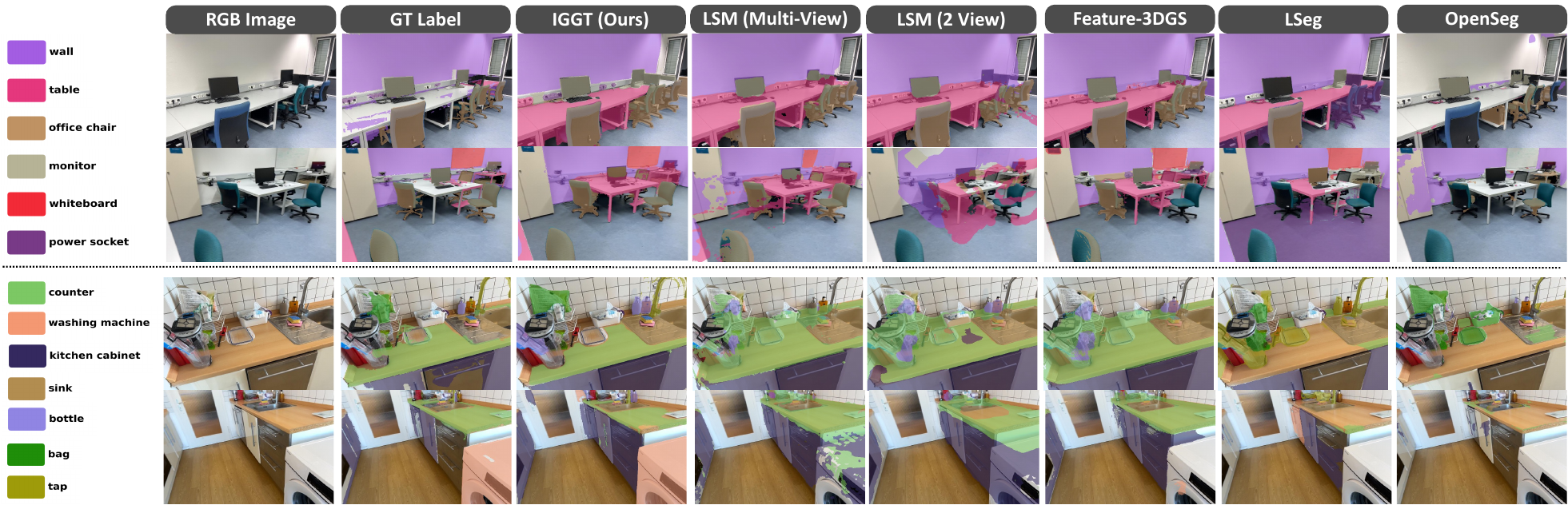}
        \caption{Qualitative Results of 2D Open-Vocabulary Segmentation on Scannet and Scannet++.}
        \label{fig:2d_ovs_vis}
        \vspace{-0.2in}
\end{figure}

\noindent \textbf{Evaluation of Open-Vocabulary Segmentation.}
We compare our method with other Image-to-3D feedforward method~\citep{fan2024large-lsm}, per-scene optimized methods~\citep{zhou2024feature,kobayashi2022decomposing}, and 2D methods~\citep{ghiasi2022scaling,lseg} on both Scannet and Scanent++ datasets. The results are reported in Tab.~\ref{tab:scannet} and Tab.~\ref{tab:scannet++}. On ScanNet++, our method achieves leading performance, surpassing other approaches by 8.34\% in mIoU for segmentation and 7.88\% in mAcc for object localization.
\begin{wrapfigure}{r}{9cm}
    \centering
    \vspace{-0.05in}
    \includegraphics[width=\linewidth]{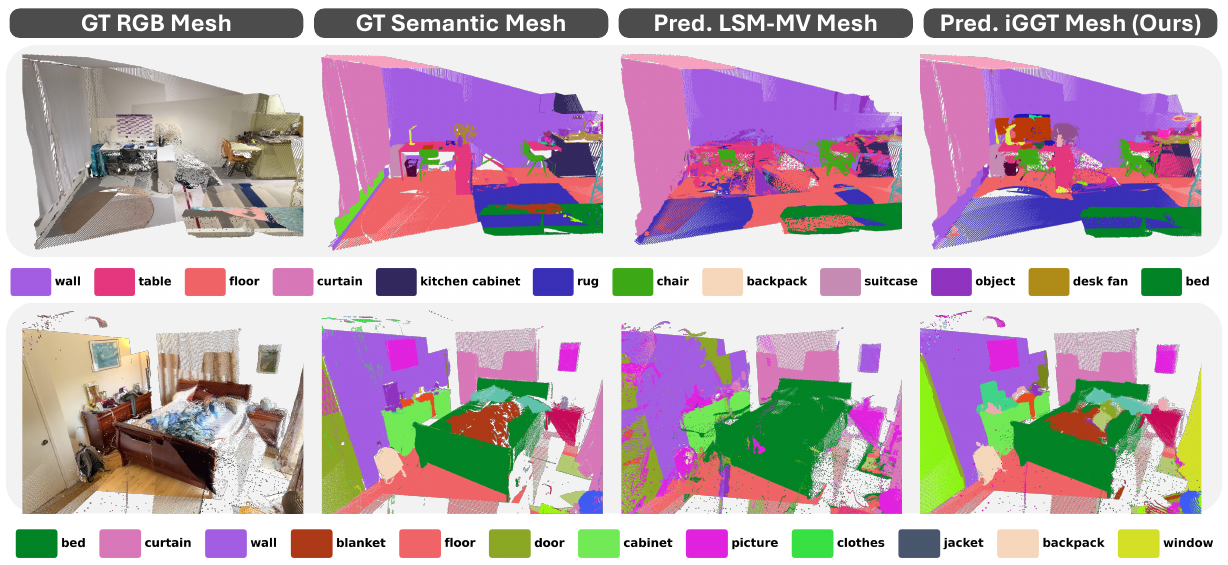}
    \caption{Visualization of 3D Open-Vocab. Segmentation.}
    \vspace{-0.15in}
    \label{fig:3dmiou}
\end{wrapfigure}
This performance improvement is attributed to our method's superior multi-view consistency, which helps correct object recognition errors caused by incomplete views, as illustrated in Fig.~\ref{fig:2d_ovs_vis}, where the sink is difficult to identify due to limited viewpoint coverage.
On the other hand, we also evaluate the accuracy of depth estimation on multi-view inputs. The results show that our method is on par with VGGT on ScanNet, and outperforms VGGT on ScanNet++ by 0.14 in Abs. Rel and 0.25 in $\tau$, benefiting from the mutual enhancement of semantics and geometry achieved through joint training.
This performance improvement is further demonstrated in 3D segmentation (see Tab.~\ref{tab:scannet} and Tab.~\ref{tab:scannet++}), where our method outperforms previous approaches by 4.31\% and 4.97\% in terms of 3D mIoU. As shown in Fig.~\ref{fig:3dmiou}, our method achieves superior 3D semantic representations while also maintaining better segmentation consistency in the same regions.

\begin{figure}[t]
    \begin{minipage}[t]{0.6\textwidth} 
        \centering
        \includegraphics[width=\linewidth]{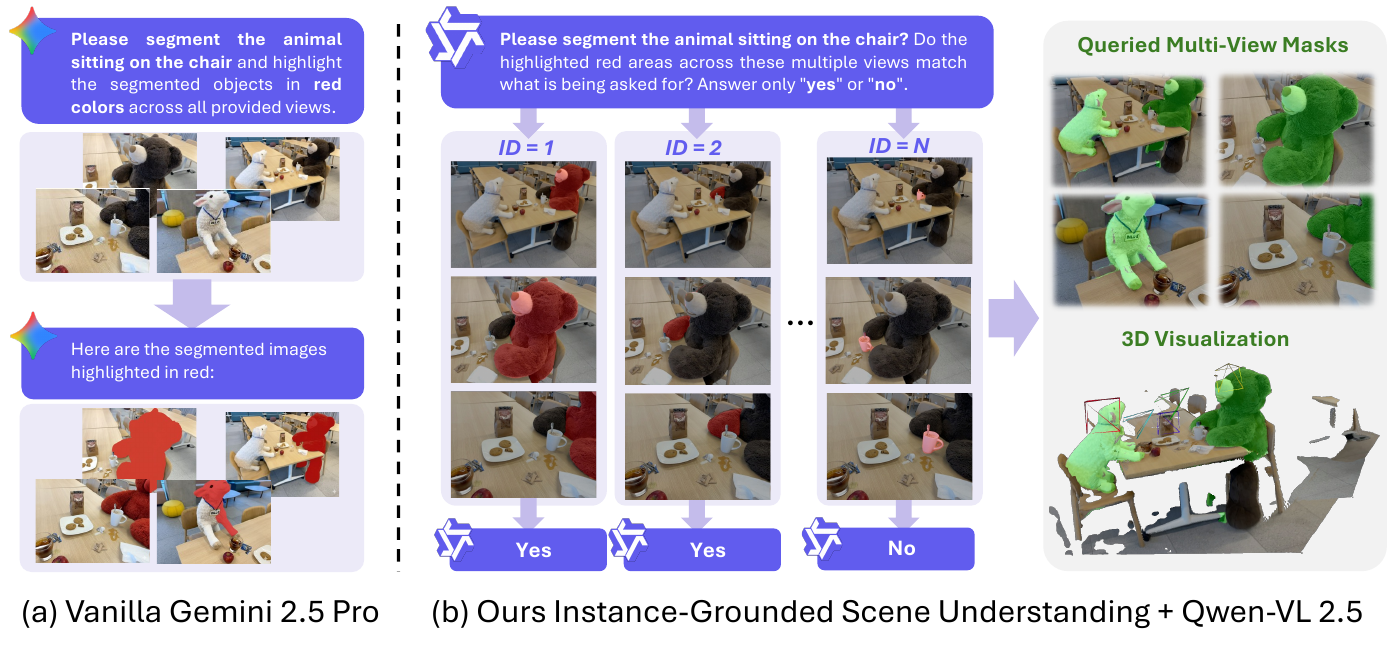}
        \vspace{-0.3in}
        \caption{Applications of QA Scene Understanding compared with vanilla Gemini 2.5 Pro~\cite{comanici2025gemini} model.}
        \label{fig:qa_mllm_standalone}
    \end{minipage}%
    \hfill 
    \begin{minipage}[t]{0.35\textwidth} 
        \centering
        \includegraphics[width=\linewidth]{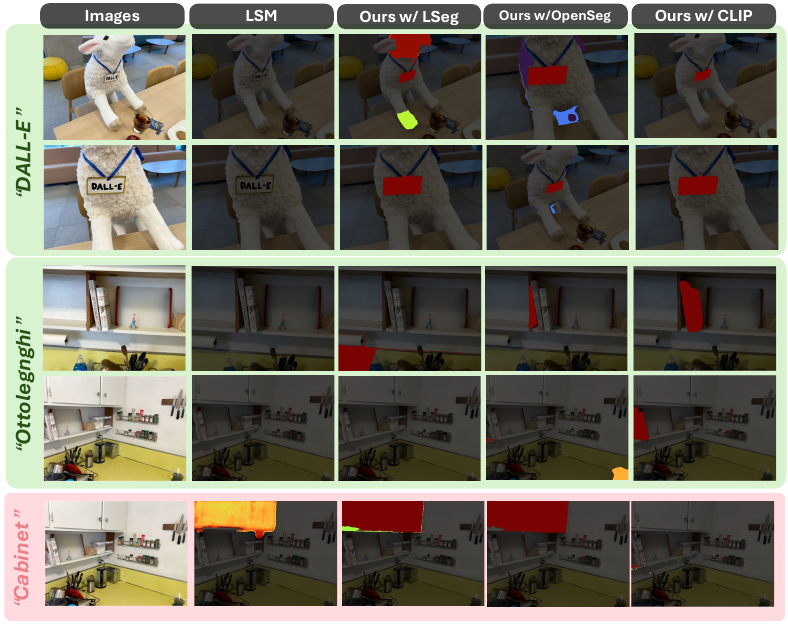}
        \vspace{-0.3in}
        \caption{Visualization of our method using different VLMs.}
        \label{fig:different_llm_standalone}
    \end{minipage}
\end{figure}

\noindent \textbf{Applications of QA Scene Grounding.}
We present the QA application results in Fig.~\ref{fig:qa_mllm_standalone} on the Teatime scene from the LERF-OVS~\citep{kerr2023lerf} dataset, and compare our approach against the state-of-the-art Gemini 2.5 Pro~\citep{comanici2025gemini}. As shown, our instance-grounded querying fully leverages the reasoning capacity of LMMs, achieving accurate segmentation for complex prompts and superior multi-view consistency compared to existing unified generation–understanding models, thereby enabling more complex QA tasks in 3D scenes.

\begin{figure}[t!]
    \vspace{-0.2in}
    \centering
    \begin{minipage}[t]{0.45\linewidth} 
        \centering
        \includegraphics[width=\linewidth]{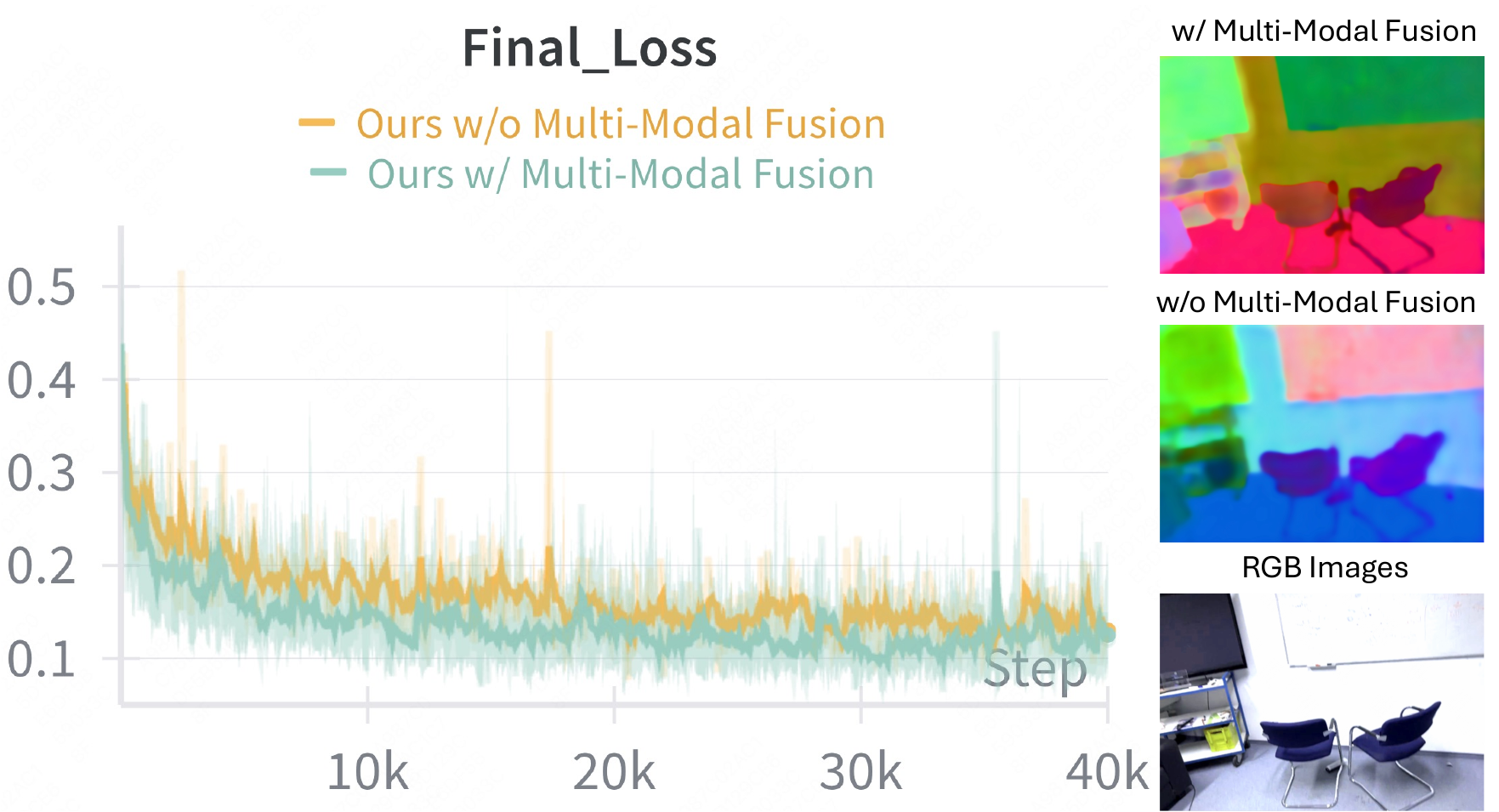}
        \captionof{figure}{Ablation on Cross-Modal Fusion.}
        \label{fig:ablation_modal_mp}
    \end{minipage}
    \hfill 
    \begin{minipage}[t]{0.53\linewidth} 
        \centering
        \vspace{-1.2in}
        \captionof{table}{Integration with Different VLMs.}
        \label{tab:ablation_2d_mp}
        \setlength{\tabcolsep}{3pt} 
        \small 
        \begin{tabular}{lcccc}
            \toprule
            \multirow{2}{*}{\textbf{Method}} & \multicolumn{2}{c}{\textbf{Scannet}} & \multicolumn{2}{c}{\textbf{Scannet++}} \\
            \cmidrule(lr){2-3} \cmidrule(lr){4-5}
            & mIoU\uparrowmetric & mAcc\uparrowmetric & mIoU\uparrowmetric & mAcc\uparrowmetric \\
            \midrule
            Ours \textit{w/} Lseg & \textbf{60.46} & \textbf{81.84} & 22.72 & 63.56 \\
            Ours \textit{w/} CLIP & 49.36 & 62.68 & 21.52 & 61.36 \\
            Ours \textit{w/} OpenSeg & 58.12 & 78.75 & \textbf{31.31} & \textbf{70.78} \\
            \bottomrule
        \end{tabular}
    \end{minipage}
    \vspace{-0.2in}
\end{figure}

\noindent \textbf{Ablation Study.}
Here, we showcase the training curve of our IGGT in Fig.~\ref{fig:ablation_modal_mp}. Without the cross-modal fusion model, the instance head struggles to capture high-resolution geometric information, resulting in more difficult convergence, as reflected in the sharpness of the chair's edges in the PCA visualization. We also conduct ablations on integrating different VLMs into our method (\textit{e.g.}, LSeg~\citep{lseg}, CLIP~\citep{radford2021clip}, OpenSeg~\citep{ghiasi2022scaling}). As shown in the table, LSeg and OpenSeg, with better global context representation, achieve higher accuracy in handling background classes (e.g., cabinet). In contrast, CLIP, with superior text alignment capabilities, performs better on complex categories, such as `DALL-E' and `Ottolegnghi' shown in Fig.~\ref{fig:different_llm_standalone}. This further demonstrates the flexibility of our method in utilizing different VLMs to achieve improved text query performance.

\subsection{Related Work}

\noindent \textbf{Spatial Foundation Models.}  
Image-to-3D reconstruction has evolved from early SfM pipelines like COLMAP~\citep{schonberger2016structure}, which estimate camera poses and sparse point clouds, to more advanced methods like 3D Gaussian Splatting (3DGS)~\citep{kerbl20233d} for efficient novel view synthesis. Scene Representation Transformers~\citep{sajjadi2022scene} represent images as latent tokens, enabling view synthesis without accurate poses, but still struggle with explicit geometry and generalization. DUSt3R~\citep{wang2024dust3r} improves upon this by directly regressing dense point maps from unposed image pairs, while VGGT~\citep{wang2025vggt} scales this approach to multiple images with competitive accuracy. However, these methods remain focused on geometric reconstruction, often neglecting higher-level scene understanding.

\noindent \textbf{3D Scene Understanding.}  
Integrating semantics into 3D reconstruction is vital for scene understanding. Methods like LangSplat~\citep{qin2024langsplat} inject vision-language features into 3D Gaussian Splatting, enabling semantic reasoning, but typically require dense multi-view inputs and per-scene optimization. Approaches such as Panst3R~\citep{zust2025panst3r} and DUSt3R~\citep{wang2024dust3r} attempt feed-forward scene understanding, but decouple geometry and semantics, limiting mutual benefits. Methods like LSM~\citep{fan2024large-lsm} and Uni3R~\citep{sun2025uni3r} align spatial models with vision-language models (e.g., LSeg~\citep{lseg}), but face limitations in integrating stronger VLMs and struggle with fine-grained, instance-level queries in complex scenes. 

\section{Conclusion}
In this paper, we introduce IGGT, a novel end-to-end framework that unifies the representation for both spatial reconstruction and contextual understanding in a 3D scene. The key to our success is that we couple geometric and instance-level semantic features by joint training and unleash the potential of a unified large transformer to achieve mutual improvements in contextual understanding and geometry reconstruction. To facilitate this task, we further present a large scale dataset called InsScene-15K, including high-quality RGB images, poses, depth maps, and 3D-consistent instance masks. Moreover, our proposed instance-grounded scene understanding strategy enables IGGT with plug-and-play integration of various VLMs and LMMs, unlocking a broader range of applications. Extensive experiments demonstrate the superiority of our IGGT over the latest state-of-the-art methods in terms of high task performance and 3D coherence. We believe that IGGT provides a promising research direction for crafting and understanding intricate 3D worlds jointly and will inspire more works in the future.

\newpage
\section{Ethics Statement}
This work focuses on improving spatial reconstruction and understanding. While our model is trained on self-annotated datasets based on standard open-source images and tested in controlled settings, we acknowledge that any AI system may potentially exhibit biases or produce unexpected behaviors. Our research is intended for academic exploration only, and we emphasize that any such outcomes do not reflect the views of the authors. We support the development of AI technologies that are ethical, safe, and aligned with societal values.

\section{Reproducibility statement}
All code and model checkpoints will be publicly released to ensure reproducibility.

\bibliography{iclr2026_conference}
\bibliographystyle{iclr2026_conference}

\newpage

\appendix
\section{Appendix}
\subsection{Use of Large Language Models}
Large Language Models (LLMs) are used exclusively for minor grammar corrections and stylistic polishing of the manuscript. They are not involved in the design of the methodology, execution of experiments, analysis of results, or any other aspect of the scientific contribution.
\subsection{Related Work}
\noindent \textbf{Spatial Foundation Model}
Image-to-3D reconstruction is a long-standing problem in computer vision. Early pipelines such as COLMAP~\citep{schonberger2016structure} and related SfM methods estimate camera poses and sparse point clouds, often followed by multi-view stereo (MVS) to obtain dense geometry~\citep{yao2018mvsnet}. Building on such SfM-based initialization, 3D Gaussian Splatting (3DGS)~\citep{kerbl20233d} introduced a highly efficient representation for photorealistic novel view synthesis, inspiring reconstruction-oriented extensions~\citep{chen2024pgsr,huang20242d}. To reduce the reliance on accurate calibration, Scene Representation Transformers~\citep{sajjadi2022scene,li2024ggrt} represent multiple images as latent scene tokens, enabling novel view synthesis under uncertain or missing poses, though they still struggle to produce explicit geometry and generalize reliably. DUSt3R~\citep{wang2024dust3r} takes a further step by directly regressing dense point maps from unposed image pairs, achieving pixel-aligned geometry without SfM initialization. In contrast, VGGT~\citep{wang2025vggt} scales this paradigm to dozens to hundreds of images in a single feed-forward pass, jointly predicting cameras, depth, point maps, and tracks with competitive accuracy to optimization-based pipelines. Despite these advances, these methods remain focused on low-level geometric reconstruction while overlooking higher-level scene understanding.

\noindent \textbf{3D Scene Understanding}
Integrating semantics into 3D reconstructions is crucial for higher-level scene understanding tasks. Recent efforts~\citep{zhou2024feature,li2024langsurf,qin2024langsplat} like LangSplat~\citep{qin2024langsplat} inject vision-language features (e.g., CLIP~\citep{radford2021clip}) into 3D Gaussian Splatting, enabling semantic reasoning over reconstructed scenes. However, these methods typically require dense multi-view inputs and per-scene optimization, which hinders scalability. More generalizable approaches like Panst3R~\citep{zust2025panst3r} build on DUSt3R~\citep{wang2024dust3r} to achieve feed-forward 3D scene understanding directly from posed or unposed images. Yet, they often decouple reconstruction from understanding and freeze the geometry module, which restricts mutual benefits between the two and leads to suboptimal semantic grounding. Parallel attempts such as LSM~\citep{fan2024large-lsm} and Uni3R~\citep{sun2025uni3r} seek to bridge geometry and semantics by aligning spatial models with specific vision-language models (e.g., LSeg~\citep{lseg}), but this tight coupling has two key drawbacks: (1) it prevents seamless integration of stronger VLMs~\citep{tschannen2025siglip, simeoni2025dinov3} as they emerge, thereby constraining text query performance; (2) the alignment is typically at the category level rather than instance-level, so these methods struggle with fine-grained, object-centric QA in scenes that contain multiple similar instances.

To address this problem, our proposed framework, IGGT, addresses these limitations by learning a unified representation for both reconstruction and understanding. Instead of tightly coupling with a single VLM, we introduce an instance-grounded paradigm where instance masks serve as a bridge to connect with diverse VLMs and Large Multimodal Models (LMMs) in a plug-and-play manner, substantially expanding downstream capabilities.

\subsection{Training Details} Our model is initialized with weights from VGGT~\citep{wang2025vggt} and fine-tuned on the InsScene-15K dataset, which contains 15{,}000 scenes. Training is performed on 8 NVIDIA A800 GPUs for 2 days using the AdamW optimizer. The learning rate is set to $1 \times 10^{-6}$ for the large unified Transformer backbone and $1 \times 10^{-5}$ for both the geometry and instance heads. For each training batch, we randomly sample 1--12 frames from a randomly selected scene, yielding a total of 24 images per batch. For hyper-parameter settings, we set \(\lambda_{pull}=2.0\), \(\lambda_{pull}=1.0\) and \(M=1.0\).

\subsection{Metrics for Different Tasks}
\begin{figure}
    \centering
    \includegraphics[width=0.75\linewidth]{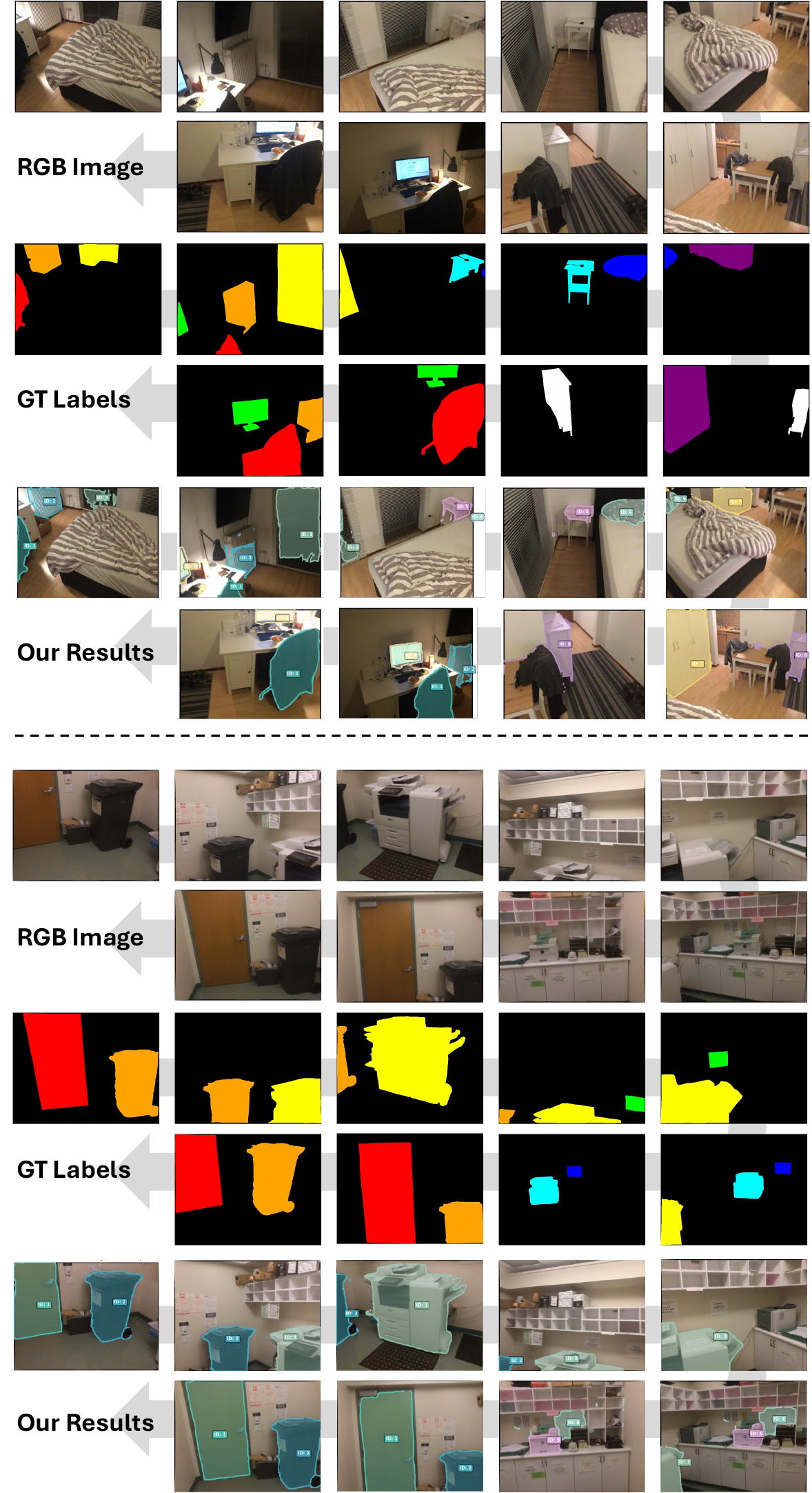}
    \caption{Visualization of our manually annotated tracking GT and our tracking results.}
    \label{fig:placeholder}
\end{figure}
\noindent \textbf{Instance Spatial Tracking}.
For the Instance Spatial Tracking task, we evaluate tracking performance using Temporal mIoU (T-mIoU) and Temporal Success Rate (T-SR). Given an object $o$ and its predicted masks $\{\hat{M}_t^o\}_{t=1}^T$ across $T$ views with corresponding ground-truth masks $\{M_t^o\}_{t=1}^T$, T-mIoU is defined as
\[
\text{T-mIoU}(o) = \frac{1}{T} \sum_{t=1}^{T} \frac{|\hat{M}_t^o \cap M_t^o|}{|\hat{M}_t^o \cup M_t^o|}.
\]
T-SR evaluates whether the object is successfully tracked across all views, and is defined as
\[
\text{T-SR}(o) = \mathbb{1}\left[\forall t \in \{1, \ldots, T\}, \; |\hat{M}_t^o| > 0 \right],
\]
where $\mathbb{1}[\cdot]$ denotes the indicator function. The final scores are averaged over all objects in the dataset.

\begin{figure}[h]
    \centering
    \includegraphics[width=1\linewidth]{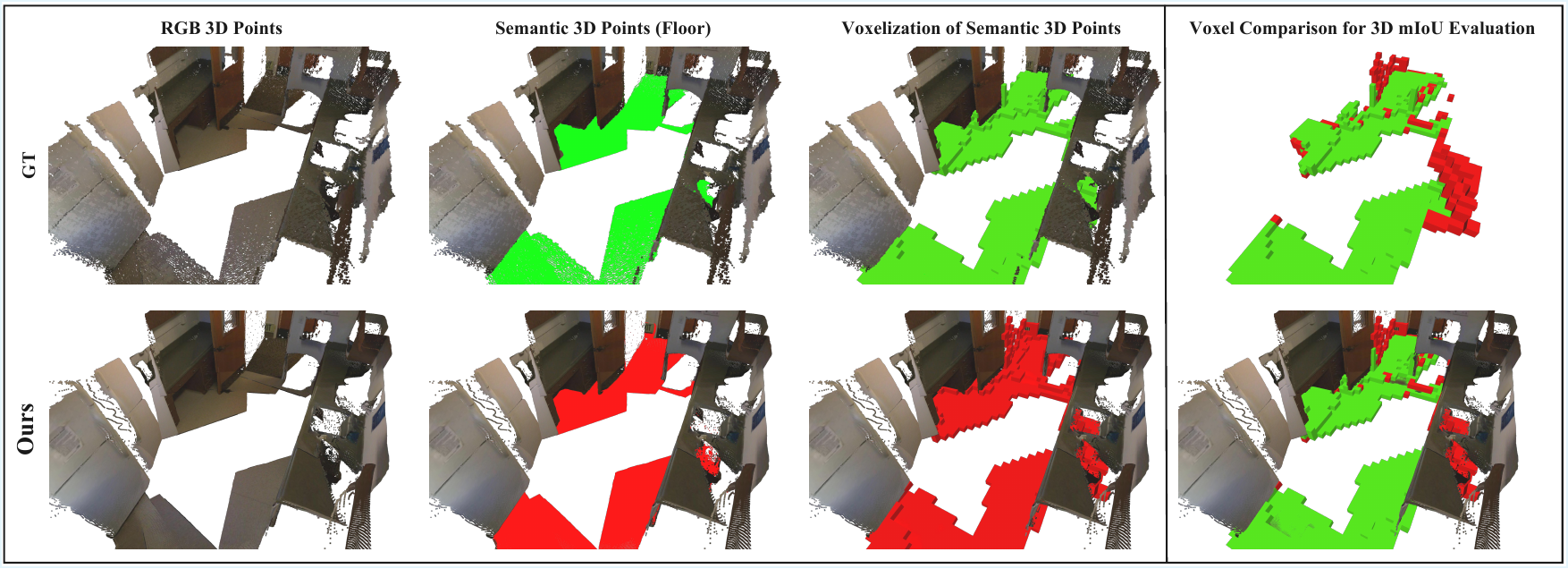}
    \caption{Visualization of the pipeline from RGB 3D points to semantic labeling, voxelization, and voxel comparison for 3D mIoU.}
    \label{fig:appendix_3dmiou_pipeline}
\end{figure}
\begin{figure}[h]
    \centering
    \includegraphics[width=1\linewidth]{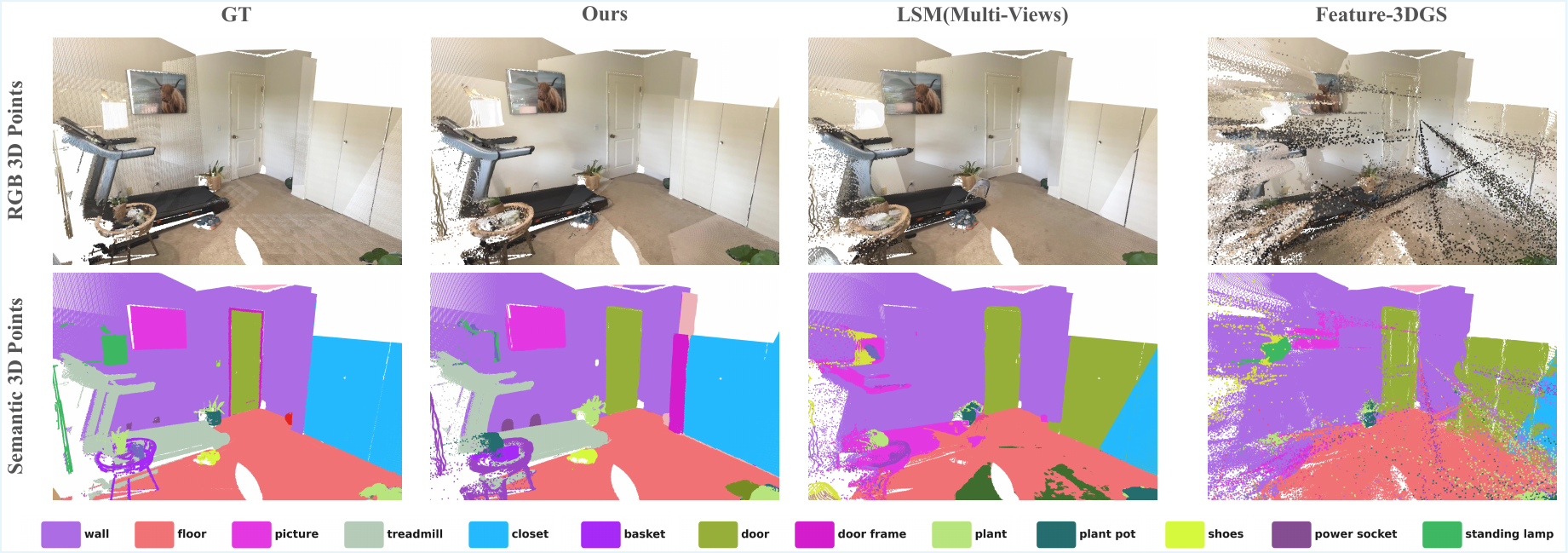}
    \caption{We visualize the RGB and semantic 3D points of the ground truth,  IGGT(Ours), LSM(Multi-Views), and Feature-3DGS.}
    \label{fig:appendix_2dvis_semantic}
\end{figure}

\textbf{3D Semantic Segmentation mIoU}. To evaluate 3D semantic segmentation, we first obtain the RGB 3D points from per-image point maps and align them with the ground truth. Next, we assign semantic labels to the corresponding 3D points based on the results of 2D open-vocabulary segmentation. These labeled 3D points are subsequently voxelized, and the 3D mIoU is computed based on the voxel representation. Fig.~\ref{fig:appendix_3dmiou_pipeline} illustrates the overall pipeline. Additionally, Fig.~\ref{fig:appendix_2dvis_semantic} presents qualitative results of 3D open-vocabulary segmentation. For LSM, based on its two-view input, we apply the global alignment strategy of Dust3R to optimize the point maps across all views. For Feature-3DGS, the ground-truth point maps are used as the initial input. However, due to the sparsity of input views, its reconstruction quality remains limited.

\subsection{Addition Visualization of our InsScene-15K dataset}
\begin{figure}
    \centering
    \includegraphics[width=1\linewidth]{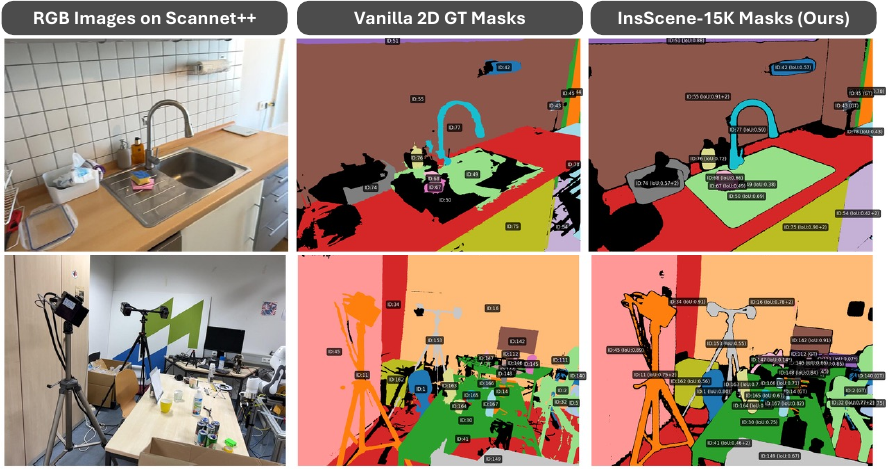}
    \caption{Comparison between vanilla Scannet++ GT masks and our refined results.}
    \label{fig:appendix_vis_scannet++}
\end{figure}

Fig.~\ref{fig:appendix_vis_scannet++} presents the vanilla masks and the refined counterparts, together with the IDs that establish the correspondence between them. The refined masks contain fewer unannotated regions and align more closely with the actual objects. Training with these high-quality instance-level masks facilitates more accurate instance-level segmentation and tracking.

\subsection{Declaration of LLM Usage}
In the preparation of this work, the authors used LLM (e.g., GPT-4) in order to improve the readability and language of the manuscript. After using this tool, the authors reviewed and edited the content as needed and take full responsibility for the content of the published article.

\subsection{Limitation} 
Our method adopts an unsupervised clustering strategy on the proposed Instance-Grounded Clustering for post-processing. As a result, the accuracy of object boundaries in the clustered masks cannot yet rival that of state-of-the-art segmentation models (e.g., SAM2~\citep{ravi2024sam2}). Future work may integrate stronger DETR-based~\citep{cheng2022masked} instance heads and larger annotated datasets to improve segmentation accuracy.

\end{document}